\documentclass[sigconf,nonacm]{aamas}

\usepackage{balance} 
\usepackage{amsmath}
\usepackage{amsthm}
\usepackage{booktabs}
\usepackage{multirow}
\usepackage{pdfpages}
\newtheorem{theorem}{Theorem}
\newtheorem{corollary}{Corollary}
\newtheorem{proposition}{Proposition} 
\DeclareMathOperator*{\argmax}{arg\,max}
\DeclareMathOperator*{\E}{\mathbb{E}}
\usepackage{xcolor}

\definecolor{gaincolor}{RGB}{25,110,70}
\definecolor{losscolor}{RGB}{165,50,50}
\newcommand{\gain}[1]{\textcolor{gaincolor}{\ensuremath{#1}}}
\newcommand{\loss}[1]{\textcolor{losscolor}{\ensuremath{#1}}}

\title{A Unified and Constrained View of Regularization-Based Robust Reinforcement Learning}

\author{Amine Andam}
\affiliation{
  \institution{Mohammed VI Polytechnic University}
  \country{Morocco}}
\email{andamamine83@gmail.com}

\author{Jamal Bentahar}
\affiliation{
  \institution{Khalifa University}
  \institution{Concordia University}
  \country{UAE}}
\email{jamal.bentahar@ku.ac.ae}
\author{Mustapha Hedabou}
\affiliation{
  \institution{Mohammed VI Polytechnic University}
  \country{Morocco}}
\email{mustapha.hedabou@um6p.ma}

\begin{abstract}
  Regularization-based methods have become a standard approach for training Deep Reinforcement Learning policies against adversarial input perturbations. In this paper, we unify these methods by deriving new upper bounds on the performance gap between the nominal and worst-case policies. Each upper bound is expressed as an existing regularization objective plus a KL-divergence penalty between the nominal and worst-case policies, which further explains why adding a KL penalty improves robustness in practice. Building on these bounds, we formulate robust training as a constrained optimization problem, showing that existing methods correspond to the special case of a fixed Lagrange multiplier. We instead update the multiplier jointly with the policy to automatically tune the regularization weight. Finally, we conduct extensive adversarial evaluations across several continuous control tasks to validate our theoretical analysis.
\end{abstract}


\newcommand{\BibTeX}{\rm B\kern-.05em{\sc i\kern-.025em b}\kern-.08em\TeX}

\begin{document}




\maketitle 
\pagestyle{plain}
\thispagestyle{plain}

\section{Introduction}
Deep Reinforcement Learning (DRL) policies that perform well can still fail under a small perturbation of their input. A bounded change to the observed state, caused by sensor noise or a deliberate adversary, can be enough to cause a large drop in their performance~\cite{DBLP:conf/nips/0001CX0LBH20}. Two families of methods address this problem. Regularization-based methods add a term to the training objective, $\mathcal{L}_{\mathrm{DRL}} + \lambda \mathcal{L}_{\mathrm{robust}}$, that encourages the policy to remain robust. Attack-driven methods instead train the policy on adversarially perturbed trajectories~\cite{DBLP:conf/iclr/ZhangCBH21,DBLP:conf/iclr/SunZLH22}. We focus on the first family, which is generally more sample efficient because it does not require collecting additional trajectories.

We consider three representative methods from this family: SA-Reg~\cite{DBLP:conf/nips/0001CX0LBH20} minimizes an upper bound on the gap between nominal and worst-case performance (i.e., $J(\pi) - J(\pi^{\mathrm{wc}})$) using a KL-divergence penalty; Radial~\cite{DBLP:conf/nips/OikarinenZMDW21} minimizes the standard DRL loss evaluated under a worst-case policy; and WocaR-RL~\cite{DBLP:conf/nips/LiangSZH22} steers the policy toward the actions with the best possible worst-case performance. A recurring empirical observation is that combining SA-Reg with either of the other two (or with attack-driven methods) consistently improves their robustness.

Even though these methods were derived from different theoretical principles, we
show that they can be unified. Our starting point is to derive, in the same spirit as SA-Reg, new upper bounds on the performance gap $J(\pi) - J(\pi^{\mathrm{wc}})$ in which the Radial and WocaR-RL regularization terms appear. Furthermore, the SA-Reg penalty appears in all new bounds, which explains why combining Radial or WocaR-RL with SA-Reg improves their robustness.

We then take a different perspective on robust training and cast it as a constrained policy optimization problem: we maximize the return subject to a lower bound on the worst-case performance under input perturbations. We use our new bounds as surrogate functions to replace the robustness constraint, providing principled approximations that can be estimated efficiently. 

This formulation also addresses the choice of the regularization weight $\lambda$. This weight controls the balance between nominal performance and robustness, making it one of the most important hyperparameters in robust training. Existing methods typically select this weight through a hyperparameter sweep and keep it fixed throughout training. Given its importance, we instead treat the choice of $\lambda$ as part of the learning problem. Our constrained formulation provides a principled way to update it: $\lambda$ increases when the robustness constraint is violated and decreases when it is satisfied.

We summarize our contributions as follows:
\begin{itemize}
    \item We unify existing regularizations by deriving new upper bounds on the performance gap between the nominal and worst-case policies. These bounds also provide a theoretical motivation for combining their regularization terms.
    \item We recast robust training as a constrained policy optimization problem. We provide tractable surrogates for the robustness constraints and update the regularization weight jointly with the policy.
    \item We complement our theoretical analysis with a thorough adversarial evaluation on several continuous control tasks. The experiments examine the robustness of combined regularizations and the proposed constrained methods against a large and diverse population of adaptive attackers.
\end{itemize}

\section{Background and Related Work}
\subsection{Preliminaries}
\subsubsection{Markov Decision Process (MDP)}

An MDP is the tuple:  \[(\mathcal{S},\mathcal{A},R,P,\rho_0,\gamma)\] where $\mathcal{S}$ is the state space, $\mathcal{A}$ is the set of actions, $R: \mathcal{S}\times \mathcal{A} \times \mathcal{S} \rightarrow  \mathbb{R} $ is the reward function, $P: \mathcal{S}\times \mathcal{A} \times \mathcal{S} \rightarrow [0,1]$ is the transition probability function, $\rho_0$ is the distribution of the initial state, and $\gamma$ is the discount factor. A policy $\pi: \mathcal{S} \rightarrow \Delta(\mathcal{A})$ is a probability distribution over the actions, and $\pi(a|s)$ denotes the probability of taking action $a$ at state $s$. In the rest of the paper, we may refer to $\pi$ as the \textit{nominal policy}.
The goal is to find a policy $\pi$ that maximizes the discounted cumulative reward $J(\pi)$:
\begin{equation}
    J(\pi) = \E_{\tau \sim \pi} \left[ \sum_{t=0}^{\infty} \gamma^t R(s_t,a_t,s_{t+1})\right]
\end{equation}

Here $\tau \sim \pi$ denotes that the trajectory $\tau = (s_0,a_0,s_1, \dots)$ is collected using policy $\pi$. Letting $R(\tau)$ denote the discounted return of a trajectory, the value function associated with policy $\pi$ is $V^{\pi}(s) =  \E_{\substack{\tau \sim \pi}} \left[ R(\tau) | s_0=s\right]$, and the action-value function is $Q^{\pi}(s,a) =  \E_{\tau \sim \pi} \left[ R(\tau)| s_0 = s, a_0 = a \right]$. The advantage function is $A^{\pi}(s,a) = Q^{\pi}(s,a) - V^{\pi}(s)$.

In many state-of-the-art policy gradient algorithms such as TRPO~\cite{DBLP:conf/icml/SchulmanLAJM15} and PPO~\cite{DBLP:journals/corr/SchulmanWDRK17}, the policy is iteratively updated by maximizing $J(\pi)$ over a local neighborhood of the current policy $\pi_k$:
\begin{equation}
    \pi_{k+1} = \arg\max_{\pi } \E_{\substack{s\sim d^{\pi_k}\\ a\sim\pi}} \left[A^{\pi_k}(s,a)\right] \quad \text{s.t.} \quad D(\pi, \pi_k) \leq \delta
    \label{eq:policy_search}
\end{equation}
where $d^\pi$ is the normalized discounted state-visitation distribution, $D$ measures the discrepancy between policies, and $\delta$ limits the size of the policy update. In practice, this expectation is estimated from a batch of trajectories collected before each update using $\pi_k$. In PPO, the explicit discrepancy constraint is replaced by a clipped surrogate objective that discourages excessively large policy updates~\cite{DBLP:journals/corr/SchulmanWDRK17}. This allows the policy to be optimized using first-order minibatch updates, avoiding the second-order computations and line search used by TRPO~\cite{DBLP:conf/icml/SchulmanLAJM15}.

\subsubsection{State Adversarial MDP (SA-MDP)} SA-MDP is a framework that formalizes the interaction between the agent and the environment in the presence of a \textit{state adversary $h: \mathcal{S} \rightarrow \mathcal{S}$} that perturbs the agent's observations: instead of receiving the original state $s_t$, it receives a perturbed state $\tilde{s}_t =h(s_t)$, and acts according to the policy $\pi(. |\tilde{s}_t)$. We consider a threat model where the adversary is constrained by an attack budget $\epsilon$ at every step: $h(s) \in \mathcal{B}_{\epsilon}(s) = \{x \in \mathcal{S}  \mid \|s-x\|_{\infty} \leq \epsilon \}$. Formally, given an MDP and the admissible perturbations $\mathcal{B}_{\epsilon}(s)$, SA-MDP is defined by the tuple $(\mathcal{S},\mathcal{A},\mathcal{B}_{\epsilon},R,P,\rho_0,\gamma)$.

We write $\pi^{\mathrm{wc}}$ to denote the worst-case policy, defined as the agent policy $\pi$ under an optimal adversary $h^*$: $\pi^{\mathrm{wc}}(.|s) = \pi(.|h^*(s))$. We denote by $J(\pi^{\mathrm{wc}})$ the discounted cumulative reward obtained under an optimal attacker $h^*$:
\begin{equation}
    \begin{aligned}
    J(\pi^{\mathrm{wc}})
    &= 
    \E_{\substack{\tau \sim \pi}} \left[ \sum_{t=0}^{\infty} \gamma^t R(s_t,a_t \sim \pi(.|h^*(s_t)), s_{t+1})\right] \\
    &= 
    \E_{\substack{\tau \sim \pi^{\mathrm{wc}}}} \left[ R(\tau)\right]
    \end{aligned}
\end{equation}

Analogous to $V^{\pi},Q^{\pi}, \text{and } A^{\pi}$, we can define worst-case RL functions ~\cite{DBLP:conf/nips/LiangSZH22,DBLP:conf/nips/0001CX0LBH20}: $V_{\mathrm{wc}}^{\pi},Q_{\mathrm{wc}}^{\pi}, \text{and } A_{\mathrm{wc}}^{\pi}$. These worst-case functions have slightly different Bellman equations; please refer to Appendix A for more details.

\subsubsection{Output bounds of policy networks~\cite{DBLP:conf/nips/XuS0WCHKLH20}} Convex relaxation techniques are used to compute certified bounds of the output of a neural network under bounded input perturbations. Given a policy network $\pi$ and a perturbation set $\mathcal{B}_{\epsilon}(s)$, they compute element-wise lower and upper bounds $\underline{\pi}$ and $\overline{\pi}$ satisfying:
\[
\underline{\pi}(.|s) \leq \pi(.|\tilde{s}) \leq \overline{\pi}(.|s)
\quad \forall \tilde{s} \in \mathcal{B}_{\epsilon}(s)
\]
These bounds guarantee that the policy output remains within the interval $[\underline{\pi},\overline{\pi}]$ for any admissible perturbation. We use the \textit{auto\_LiRPA} library to compute these bounds ~\cite{DBLP:conf/nips/XuS0WCHKLH20}.

\subsubsection{Policy Performance Bounds} In the following, we state a theorem from ~\cite{DBLP:conf/icml/AchiamHTA17} that is used throughout our paper to provide the new robustness bounds.
\begin{theorem} 
\label{achiam_theorem}
\cite{DBLP:conf/icml/AchiamHTA17}
        
For any function $f:\mathcal{S}\to\mathbb{R}$ and any policies $\pi'$ and $\pi$, define \[\delta_f(s,a,s') \doteq R(s,a,s')+\gamma f(s')-f(s)\]
\[\epsilon_f^{\pi'} \doteq \max_s \left| \mathbb{E}_{a\sim\pi',\,s'\sim P}\!\left[\delta_f(s,a,s')\right]\right|\] and: 
\[
    L_{\pi,f}(\pi') \doteq
    \mathbb{E}_{\substack{s\sim d^\pi\\ a\sim\pi\\ s'\sim P}}
    \left[\left(\frac{\pi'(a|s)}{\pi(a|s)}-1\right)\delta_f(s,a,s')\right]
\]    
\[
    D_{\pi,f}^{\pm}(\pi') \doteq
    \frac{L_{\pi,f}(\pi')}{1-\gamma}
    \pm
    \frac{2\gamma\epsilon_f^{\pi'}}{(1-\gamma)^2}
    \mathbb{E}_{s\sim d^\pi}\!\left[D_{TV}(\pi\|\pi')[s]\right]
\]

\noindent Here $D_{TV}$ is the total variation divergence. 

The following bounds hold:
\begin{equation}
    D_{\pi,f}^{-}(\pi')
    \le
    J(\pi')-J(\pi)
    \le
    D_{\pi,f}^{+}(\pi')
\end{equation}
\end{theorem}

\subsubsection{Constrained policy optimization}  In addition to finding policies that maximize returns, we are often interested in policies that do so while satisfying a set of constraints. In these settings, we usually augment the MDP with a set of cost functions $C: \mathcal{S}\times \mathcal{A} \times \mathcal{S} \rightarrow  \mathbb{R}$ and limits $d$. Specifically, we look for policies that maximize the return $J(\pi)$, while keeping the cumulative costs $J_{\mathrm{C}}(\pi) = \E_{\substack{\tau \sim \pi}} \left[\sum_{t=0}^{\infty} \gamma^t C(s_t,a_t,s_{t+1})\right]$ below the limit $d$. Formally, we write ~\cite{DBLP:conf/icml/AchiamHTA17}:
\begin{equation}
\label{eq:general_constr}
\begin{aligned}    
        \pi_{k+1} 
        &= 
        \argmax_{\pi } J(\pi) \\
        & \text{s.t.}\quad J_{\mathrm{C}}(\pi) \le d 
\end{aligned}
\end{equation}
Lagrangian relaxation techniques \cite{Achiam2019BenchmarkingSE,DBLP:conf/iclr/TesslerMM19} are often used to solve these optimization problems. They consist in transforming the constrained problem Eq. (\ref{eq:general_constr}) into the following equivalent unconstrained problem:
\begin{equation}
\label{eq:lag_form}
    \max_{\pi}\min_{\lambda \ge 0} \left[J(\pi) -\lambda \left(J_{\mathrm{C}}(\pi) - d\right)\right]
\end{equation}

\noindent where $\lambda \ge 0$ is the Lagrange multiplier. The problem in  Eq. (\ref{eq:lag_form}) is solved in an alternating style: a gradient ascent to optimize policy $\pi$, followed by a gradient descent to update $\lambda$. The value of $\lambda$ tends to increase when the constraint is violated and vice versa. 

\subsection{Related Work}
\label{sec:background}

Algorithms for training policies that are robust to input perturbations generally fall into two categories. \textbf{Regularization-based} methods add a robustness term to the training objective (i.e., $\mathcal{L}_{\mathrm{DRL}} + \lambda \mathcal{L}_{\mathrm{robust}}$). \textbf{Attack-driven} methods instead train the policy on adversarially perturbed trajectories: \[\tilde{\tau} = (\dots,h^*(s_t),\pi(h^*(s_t)),h^*(s_{t+1}), \dots)\] where the main challenge is to design an optimal adversary $h^*$. Such an adversary is typically learned using DRL, with the objective of minimizing the victim's return \cite{DBLP:conf/iclr/ZhangCBH21,DBLP:conf/iclr/SunZLH22}, which requires collecting additional trajectories. Regularization-based approaches avoid this requirement and are therefore generally preferable. In this work, we focus exclusively on regularization-based methods and review the main state-of-the-art approaches in this category below.

\paragraph{SA-Reg~\cite{DBLP:conf/nips/0001CX0LBH20}.} The idea behind the State-Adversarial Regularizer (SA-Reg) is to keep the worst-case performance close to the nominal one by bounding the gap between them. To do so, the authors propose the following upper bound: 
\begin{equation}
    \label{eq:orig_sa}
    J(\pi) - J(\pi^\mathrm{wc}) \leq \tilde{\alpha}^{R} \max_{s} D_{\mathrm{TV}}\left(\pi\|\pi^\mathrm{wc}\right)[s]
\end{equation}
where $\tilde{\alpha}^{R}=2 \left[1+\frac{\gamma}{(1-\gamma)^2}\right]\max_{s,a,s'} |R(s,a,s')|$. 

Implementation-wise, they regularize the policy by minimizing the following loss:
\begin{equation}
    \mathcal{L}^{\mathrm{SA-Reg}} = \E_{s \sim d^{\pi}} \left[D_{\mathrm{TV}}\left(\pi\| \pi^\mathrm{wc}\right)[s]\right]
\end{equation}

\noindent where the divergence term $D_{\mathrm{TV}}\left(\pi\| \pi^\mathrm{wc}\right)$ is computed using convex relaxation tools as:
\begin{equation}
    D_{\mathrm{TV}}(\pi\|\pi^{\mathrm{wc}})[s] = \max_{\tilde{s} \in \mathcal{B}_{\epsilon}(s)} D_{\mathrm{TV}}\left(\pi(.|s)\|\pi(.|\tilde{s})\right)
\end{equation}

\paragraph{Radial~\cite{DBLP:conf/nips/OikarinenZMDW21}.} The motivation behind Radial is to minimize an upper bound of the standard DRL loss when evaluated on adversarial states: 
\begin{equation}
    \mathcal{L}^{\mathrm{DRL}}(\tilde{s}) \leq \mathcal{L}^{\mathrm{Radial}}(s) \quad  \forall s \in \mathcal{S}, \forall \tilde{s} \in \mathcal{B}_{\epsilon}(s)
\end{equation}
For PPO, they suggest the following regularization:
\begin{equation}
\label{eq:old_radial}
    \mathcal{L}^\mathrm{Radial} = \mathbb{E}_{t}\left[ 
        -\min \left( r^{\mathrm{wc}}_t A_t, 
        \mathrm{clip}(r^{\mathrm{wc}}_t, 1 \pm \epsilon_{\mathrm{ppo}}) A_t
        \right) \right]
\end{equation}
where $r_t^{\mathrm{wc}} = \frac{\pi^{\mathrm{wc}}(a_t \mid s_t)}{\pi_{{\text{old}}}(a_t \mid s_t)}$ and $\epsilon_{\mathrm{ppo}}$ is the PPO clipping parameter. 

Radial gives a closed form for $\pi^{\mathrm{wc}}$ using convex relaxation tools: it assigns the lowest possible probability to better-than-average actions $A^{\pi}(s,a) \geq 0$, and the highest possible probability to worse-than-average actions $A^{\pi}(s,a) < 0$:
\begin{equation}
\label{eq:radial_wc}
    \pi^{\mathrm{wc}}(a|s)= \mathbb{I}_{A^\pi(s,a)\ge0}\,\underline{\pi}(a|s) + \mathbb{I}_{A^\pi(s,a)<0}\,\overline{\pi}(a|s)
\end{equation}

\paragraph{WocaR-RL~\cite{DBLP:conf/nips/LiangSZH22}.} 
WocaR-RL optimizes the policy to select actions that maximize not only the nominal performance but also the worst-case performance. Analogous to how the advantage $A^\pi$ is used in PPO to encourage high-return actions, WocaR-RL uses the worst-case action-value function $Q_{\mathrm{wc}}^\pi$ to encourage actions that remain good under attack. They propose the following loss: 
\begin{equation}
\label{eq:orig_wocar}
\mathcal{L}^\mathrm{WocaR-RL} = \mathbb{E}_t \Big[ -\min \big( r^\pi_t Q^{\pi}_{\mathrm{wc}},\mathrm{clip}(r^\pi_t, 1 \pm \epsilon_{\mathrm{ppo}})  Q^{\pi}_{\mathrm{wc}}\big) \Big]
\end{equation}
where $r_t^{\pi} = \frac{\pi(a_t|s_t)}{\pi_{\mathrm{old}}(a_t|s_t)}$.
In addition, they propose an efficient way to train $Q^{\pi}_{\mathrm{wc}}$ without collecting adversarial trajectories (see Appendix A, Definition 3).

\section{Unifying Robust Regularization}
In this section, we recast the three regularization methods presented in Section~\ref{sec:background} as upper bounds on the performance gap between the nominal and worst-case policies (i.e., $J(\pi) - J(\pi^{\mathrm{wc}})$). We show that each bound can be estimated from the batch already collected under $\pi_k$ (i.e., $s \sim d^{\pi_k}$). Proofs are deferred to Appendix B.

\subsection{SA-Reg}
We revisit the bound provided by SA-Reg~\cite{DBLP:conf/nips/0001CX0LBH20} in two ways. First, the original bound in Eq.~(\ref{eq:orig_sa}) uses an incorrect constant $\tilde{\alpha}^R$, which we correct. Second, we replace the maximum over states with an expectation over $d^\pi$, matching how SA-Reg is implemented.
\begin{theorem}
\label{theorem_new_sa} 
For any policy $\pi$ and its worst-case counterpart $\pi^{\mathrm{wc}}$, the following inequality holds:
\begin{equation}
    \label{eq:new_sa_reg}
        J(\pi) - J(\pi^{\mathrm{wc}})  \leq \alpha^R   \E_{s\sim d^\pi}\!\left[D_{TV}(\pi\|\pi^{\mathrm{wc}})[s]\right]
\end{equation}
\noindent where $\alpha^R = \frac{2\max_{s,a,s'} \left|R(s,a,s')\right| }{(1-\gamma)^2}$.
\end{theorem}

The upper bound in Theorem~\ref{theorem_new_sa} depends on $d^\pi$, whereas the available batch is collected under $\pi_k$. To obtain a bound in terms of $d^{\pi_k}$, we use the following proposition.

\begin{proposition}
\label{prop:sampled_return_difference}
For any policies $\pi$ and $\pi_k$, the following inequality holds:
\begin{equation}
\label{sup:sa_reg_2}
J(\pi)-J(\pi_k)
\leq
\alpha^R
\E_{s\sim d^{\pi_k}}
\!\left[D_{\mathrm{TV}}(\pi_k\|\pi)[s]\right]
\end{equation}
\end{proposition}

Using Proposition~\ref{prop:sampled_return_difference} and the triangle inequality for $D_{\mathrm{TV}}$, we obtain the following corollary:

\begin{corollary}
\label{cor-sa-reg}
For any policies $\pi$ and $\pi_k$, with $\pi^{\mathrm{wc}}$ denoting the worst-case counterpart of $\pi$, the following inequality holds:
\begin{equation}
\label{eq:new_sa_reg_rigo}
\begin{aligned}
&J(\pi)-J(\pi^{\mathrm{wc}})\\
&\le
\alpha^R \E_{s\sim d^{\pi_k}}
\left[D_{\mathrm{TV}}(\pi\|\pi^{\mathrm{wc}})[s]\right] + 2\alpha^R \E_{s\sim d^{\pi_k}}
\left[D_{\mathrm{TV}}(\pi\|\pi_k)[s]\right]
\end{aligned}
\end{equation}
\end{corollary}
We include Eq.~(\ref{eq:new_sa_reg_rigo}) for theoretical completeness. It provides a rigorous justification for using SA-Reg with samples collected under the current policy $\pi_k$. The second term of this upper bound is not optimized explicitly, and it is implicitly controlled when using policy optimization algorithms such as PPO.

\subsection{Radial Inspired Bound}

The original Radial loss~\cite{DBLP:conf/nips/OikarinenZMDW21},given in Eq.~(\ref{eq:old_radial}), is derived specifically for PPO and relies on the particular worst-case policy in Eq. (\ref{eq:radial_wc}). We derive a new upper bound that (i) casts the Radial objective as minimizing a performance gap, as we did for SA-Reg, and (ii) extends it to the general policy-optimization framework of Eq. (\ref{eq:policy_search}) and to any worst-case policy.

First, Proposition~\ref{proposition_new_radial} rewrites the (unclipped) Radial objective as an expectation of the advantage under any valid worst-case policy.

\begin{proposition}
\label{proposition_new_radial}
For any worst-case policy $\pi^{\mathrm{wc}}$, the Radial regularization term can be written as:
\begin{equation}
\mathcal{L}^\mathrm{Radial} = - \E_{\substack{s \sim d^{\pi_k}\\ a \sim \pi^{\mathrm{wc}}}}\left[A^{\pi_k}(s,a)\right]
\end{equation}
\end{proposition}

Next, we derive an upper bound on the performance gap that contains the Radial loss.
\begin{theorem}
\label{theorem_new_radial}
For any policy $\pi$ and its worst-case counterpart $\pi^{\mathrm{wc}}$, the following inequality holds:
\begin{equation}
\begin{aligned}
    &J(\pi) - J(\pi^{\mathrm{wc}}) \\
        &\leq \frac{-1}{1-\gamma} \E_{\substack{s\sim d^\pi\\ a\sim\pi^{\mathrm{wc}}}} \left[A^{\pi}(s,a)\right] + \alpha^A \E_{s\sim d^\pi}\!\left[D_{TV}(\pi\|\pi^{\mathrm{wc}})[s]\right] 
\end{aligned}
\end{equation}
\noindent where $\alpha^A = \frac{2\gamma  \max_{s,a} \left| A^{\pi}(s,a)\right|}{(1-\gamma)^2}$.
\end{theorem}

The first term in Theorem~\ref{theorem_new_radial} is the Radial objective from Proposition~\ref{proposition_new_radial}, and the second is the SA-Reg penalty. Thus, this decomposition explains why combining the two regularizations improves robustness in practice. Unlike the original derivation, our bound holds for any worst-case policy, not only the construction in Eq. (\ref{eq:radial_wc}).

As for SA-Reg, using Proposition~\ref{prop:sampled_return_difference} and the triangle inequality for $D_{\mathrm{TV}}$, we obtain the following bound in terms of $d^{\pi_k}$:
\begin{corollary}
\label{cor-radial}
For any policies $\pi$ and $\pi_k$, with $\pi^{\mathrm{wc}}$ denoting the worst-case counterpart of $\pi$, the following inequality holds:
\begin{equation}
\begin{aligned}
    &J(\pi) - J(\pi^{\mathrm{wc}})    \\
    &\le \frac{-1}{1-\gamma} \E_{\substack{s\sim d^{\pi_k}\\ a\sim\pi^{\mathrm{wc}}}} \left[A^{\pi_k}(s,a)\right] + \alpha^A \E_{s\sim d^{\pi_k}}\!\left[D_{TV}(\pi\|\pi^{\mathrm{wc}})[s]\right]\\
    & \qquad + \alpha^{AR} \E_{s\sim d^{\pi_k}}\!\left[D_{TV}(\pi\|\pi_k)[s]\right] 
\end{aligned}
\end{equation}
where $\alpha^{AR} = \alpha^A + \alpha^R$.
\end{corollary}
\subsection{WocaR-RL Inspired Bound}
We revisit the regularization of WocaR-RL. We first reformulate it using the worst-case advantage function, which is the term that naturally appears in the performance bounds. We then derive a new upper bound that provides a theoretical justification for combining WocaR-RL with the SA-Reg regularizer.

\begin{proposition}
\label{proposition_new_wocar}
The unclipped WocaR-RL regularization can be written as:
\begin{equation}
    \label{eq_proposition_new_wocar}
    \mathcal{L}^\mathrm{WocaR-RL} = - \E_{\substack{s \sim d^{\pi_k}\\ a \sim \pi}}\left[A^{\pi_k}_{\mathrm{wc}}(s,a)\right]
\end{equation}
\end{proposition}

Although the original algorithm is formulated using the worst-case action-value function, replacing it with the worst-case advantage does not change the optimization problem since:
\[
\nabla_{\theta}
\mathbb{E}_{\substack{s\sim d^{\pi_k}\\a\sim\pi_\theta}}
\!\left[A_{\mathrm{wc}}^{\pi_k}(s,a)\right]
=
\nabla_{\theta}
\mathbb{E}_{\substack{s\sim d^{\pi_k}\\a\sim\pi_\theta}}
\!\left[Q_{\mathrm{wc}}^{\pi_k}(s,a)\right]
\]

The following theorem shows that the reformulated WocaR-RL objective upper bounds the robust performance gap up to the same regularization term used in SA-Reg.

\begin{theorem}
\label{theorem_new_wocar}
For any policy $\pi$ and its worst-case counterpart $\pi^{\mathrm{wc}}$, the following inequality holds:
\begin{equation}
\begin{aligned}
&J(\pi)-J(\pi^{\mathrm{wc}}) \\
&\le
-\frac{1}{1-\gamma}\E_{\substack{s\sim d^\pi\\ a\sim\pi}} \!\left[A_{\mathrm{wc}}^{\pi}(s,a)\right] + 2 \alpha^R \E_{s\sim d^\pi} \!\left[D_{TV}(\pi\|\pi^{\mathrm{wc}})[s]\right]
\end{aligned}
\end{equation}
\end{theorem}

As with the previous bounds, Proposition~\ref{cor-wocar} expresses the state expectations under $d^{\pi_k}$, allowing estimation from the existing batch. Its proof is provided in the appendix.

\begin{proposition}
\label{cor-wocar}
For any policies $\pi$ and $\pi_k$, with $\pi^{\mathrm{wc}}$ denoting the worst-case counterpart of $\pi$, the following inequality holds:
\begin{equation}
\begin{aligned}
&J(\pi)-J(\pi^{\mathrm{wc}}) \\
&\le\; -\frac{1}{1-\gamma} \E_{\substack{s\sim d^{\pi_k}\\ a\sim\pi}} \!\left[A_{\mathrm{wc}}^{\pi}(s,a)\right]  + 2\alpha^R \E_{s\sim d^{\pi_k}} \!\left[D_{TV}(\pi\|\pi^{\mathrm{wc}})[s]\right]  \\
    &\qquad + 5\alpha^R \E_{s\sim d^{\pi_k}} \!\left[D_{TV}(\pi\|\pi_k)[s]\right]
\end{aligned}
\end{equation}
\end{proposition}

\section{Robustness as a Constraint}
\label{sec:constrainted}
Regularization-based methods take the general form of Eq.~(\ref{eq:general_reg}), where the regularization weight $\lambda$ is fixed throughout training. This weight is among the most important hyperparameters for training robust agents, and is typically found through a hyperparameter sweep. In this section, we propose an alternative formulation that updates $\lambda$ automatically, increasing it when the policy is not yet robust enough and decreasing it as robustness improves.
\begin{equation}
\label{eq:general_reg}
\min_{\pi}\mathcal{L}_{\mathrm{DRL}}(\pi) + \lambda \mathcal{L}_{\mathrm{robust}}(\pi)    
\end{equation}

To this end, we first notice that regularized training in Eq.~(\ref{eq:general_reg}) can be seen as a Lagrangian relaxation with a fixed Lagrange multiplier (see Eq.~(\ref{eq:lag_form})). We propose instead to formalize the robust training as a constrained problem solved using Lagrangian relaxation with a \emph{learned} $\lambda$, which allows us to have an adaptive regularization weight. The new DRL problem is a policy optimization with a constraint on the policy robustness. We consider the following constrained problem:
\begin{equation}
\label{eq:constrained_robustness}
\begin{aligned}    
\pi_{k+1} 
    &= 
    \argmax_{\pi } J(\pi) \\
    & \text{s.t.}\quad J(\pi^{\mathrm{wc}}) \geq d \quad \text{and} \quad D(\pi,\pi_k) \leq \delta
\end{aligned}
\end{equation}

\noindent where the constraint is on the worst-case performance exceeding a predefined limit $d$. Next, we explore different adaptations of the worst-case constraint. We refer to the proposed family of methods as \emph{RaC} (Robustness as a Constraint). To avoid notational clutter, we omit the trust-region constraint $D(\pi,\pi_k) \leq \delta$ in the following discussion.  
\subsection{Worst-Case Monotonic Improvement}
\label{sec:cpo_mono}
A reasonable constraint is to aim for monotonic improvement of the worst-case performance, which is equivalent to selecting $d=J(\pi_k^{\mathrm{wc}})$. In this case, we obtain the following constrained problem:
\begin{equation}
\label{eq:monotonic_improv}
\begin{aligned}    
    \pi_{k+1} 
        &= 
        \argmax_{\pi } J(\pi) \\
        & \text{s.t.}\quad J(\pi^{\mathrm{wc}}) \geq J(\pi_k^{\mathrm{wc}})
\end{aligned}
\end{equation}
To evaluate the monotonic constraint, we first derive a lower bound on $J(\pi^{\mathrm{wc}}) - J(\pi_k^{\mathrm{wc}})$ in Theorem \ref{theorem_monotonic} that can be used as an easy-to-estimate surrogate.

\begin{theorem}
\label{theorem_monotonic}
For any policy $\pi$, the following inequality holds:
\begin{equation}
    \begin{aligned}
        &J(\pi^{\mathrm{wc}}) - J(\pi^{\mathrm{wc}}_k) \\
        &\geq \frac{1}{1-\gamma}\mathbb{E}_{\substack{s\sim d^{\pi_k}\\ a\sim \pi}} \left[A_{\mathrm{wc}}^{\pi_k}(s,a)\right] - \alpha^A_{\mathrm{wc}} \E_{s\sim d^{\pi_k}}\left[D_{TV}(\pi\|\pi^{\mathrm{wc}})[s]\right] \\
        & \qquad -\gamma \alpha^A_{\mathrm{wc}}  \E_{s\sim d^{\pi_k}}\!\left[D_{TV}(\pi\|\pi_k)[s]\right]
    \end{aligned}
\end{equation}
\noindent where $\alpha^A_{\mathrm{wc}} = \frac{2 \max_{s,a}\left|A_{\mathrm{wc}}^{\pi_k}(s,a) \right|}{(1-\gamma)^2}$.
\end{theorem}

Let $\mathcal{C}(\pi)$ denote the right-hand side of the bound in Theorem~\ref{theorem_monotonic}:
\begin{equation}
    \begin{aligned} 
        &\mathcal{C}(\pi) = \frac{1}{1-\gamma}\mathbb{E}_{\substack{s\sim d^{\pi_k}\\ a\sim \pi}} \left[A_{\mathrm{wc}}^{\pi_k}(s,a)\right] - \alpha^A_{\mathrm{wc}} \E_{s\sim d^{\pi_k}}\left[D_{TV}(\pi\|\pi^{\mathrm{wc}})[s]\right] \\
        & \qquad \qquad -\gamma \alpha^A_{\mathrm{wc}}  \E_{s\sim d^{\pi_k}}\!\left[D_{TV}(\pi\|\pi_k)[s]\right]
    \end{aligned}
\end{equation}
Imposing $\mathcal{C}(\pi)\geq 0$ is sufficient to ensure $J(\pi^{\mathrm{wc}})\geq J(\pi_k^{\mathrm{wc}})$. We optimize the resulting constrained problem using Lagrangian relaxation:
\begin{equation}
    \label{eq:rpc_mono_general}
    \max_{\pi}\min_{\lambda \ge 0} \left[J(\pi) + \lambda  \mathcal{C}(\pi)\right]
\end{equation}

The policy is updated by gradient ascent:
\[
    \theta \leftarrow \theta + \nu_1 \left( \nabla_\theta J+\lambda\nabla_\theta\mathcal{C}\right)
\]

while the multiplier is updated by gradient descent
subject to $\lambda\geq 0$:
\[ 
    \lambda \leftarrow \left[\lambda - \nu_2 \mathcal{C}\right]_{+}
\]
where $\nu_1$ and $\nu_2$ are learning rates and $[x]_+=\max\{0,x\}$. We refer to this method as \emph{RaC-Mono}. \emph{RaC-Mono} is closely related to WocaR-RL through its use of the worst-case advantage. It also includes an explicit SA-Reg-style divergence penalty between the nominal and worst-case policies. Given this connection to WocaR-RL, we focus the gap-tolerance variants below on SA-Reg and Radial rather than introducing a separate RaC-WocaR variant.
\subsection{Gap Tolerance}
Another plausible constraint is to impose a maximum tolerable gap between the nominal and worst-case returns, which is equivalent to
selecting $d = J(\pi) - \Delta_{\mathrm{max}}$, where $\Delta_{\mathrm{max}}$ is a threshold. 

Therefore, we obtain the following constrained problem:
\begin{equation}
\label{eq:constrained_gap}
\begin{aligned}    
\pi_{k+1} 
    &= 
    \argmax_{\pi } J(\pi) \\
    & \text{s.t.}\quad J(\pi) -J(\pi^{\mathrm{wc}}) \leq \Delta_{\mathrm{max}}
\end{aligned}
\end{equation}

This new constraint raises two practical challenges: (1) choosing $\Delta_{\mathrm{max}}$ and (2) estimating the performance gap $J(\pi) -J(\pi^{\mathrm{wc}})$. We propose the following choice for $\Delta_{\mathrm{max}}$:
\begin{equation}
    \Delta_{\mathrm{max}} = \eta \times \left|J(\pi_k)\right| \quad \eta \in (0,1]
\end{equation}
With this choice, $\Delta_{\mathrm{max}}$ represents the largest degradation we are willing to accept under attack, expressed as a fraction $\eta$ of the nominal performance. For example, $\eta = 0.1$ allows the return to degrade by at most $10\%$ of $|J(\pi_k)|$.

For the second challenge, we recall that Corollaries~\ref{cor-sa-reg} and~\ref{cor-radial} provide tractable upper bounds on the performance gap. Therefore, constraining any of these upper bounds to be below $\Delta_{\mathrm{max}}$ is sufficient to guarantee that the original constraint is satisfied. We propose two new methods \emph{RaC-SA} and \emph{RaC-Radial} that use Corollaries~\ref{cor-sa-reg} and \ref{cor-radial}, respectively. They both use the following Lagrangian formulation:
\begin{equation}
    \label{eq:cpo_general}
    \max_{\pi}\min_{\lambda \ge 0} \left[J(\pi) -\lambda \left( \mathcal{C}(\pi) - \Delta_{\mathrm{max}}\right)\right]
\end{equation}
For \emph{RaC-SA}:
\begin{equation}
    \mathcal{C}(\pi) = \alpha^R \E_{s\sim d^{\pi_k}}
    \left[D_{\mathrm{TV}}(\pi\|\pi^{\mathrm{wc}})[s]\right] + 2\alpha^R \E_{s\sim d^{\pi_k}}
    \left[D_{\mathrm{TV}}(\pi\|\pi_k)[s]\right]
\end{equation}
and for \emph{RaC-Radial}:
\begin{equation}
    \begin{aligned}
    &\mathcal{C}(\pi) = \frac{-1}{1-\gamma} \E_{\substack{s\sim d^{\pi_k}\\ a\sim\pi^{\mathrm{wc}}}} \left[A^{\pi_k}(s,a)\right] + \alpha^A \E_{s\sim d^{\pi_k}}\!\left[D_{TV}(\pi\|\pi^{\mathrm{wc}})[s]\right]\\
    & \qquad + \alpha^{AR} \E_{s\sim d^{\pi_k}}\!\left[D_{TV}(\pi\|\pi_k)[s]\right] 
\end{aligned}
\end{equation}

The policy is updated by gradient ascent:
\[
    \theta \leftarrow \theta + \nu_1
    \left(\nabla_\theta J-\lambda\nabla_\theta\mathcal{C}\right).
\]
The multiplier is updated by gradient descent subject to $\lambda\geq 0$:
\[
    \lambda \leftarrow
    \left[\lambda+\nu_2
    \left(\mathcal{C}(\pi)-\Delta_{\mathrm{max}}\right)\right]_{+}
\]

\subsection{Practical implementation}
\begin{figure*}[t]
    \centering
    \includegraphics[width=0.85\textwidth]{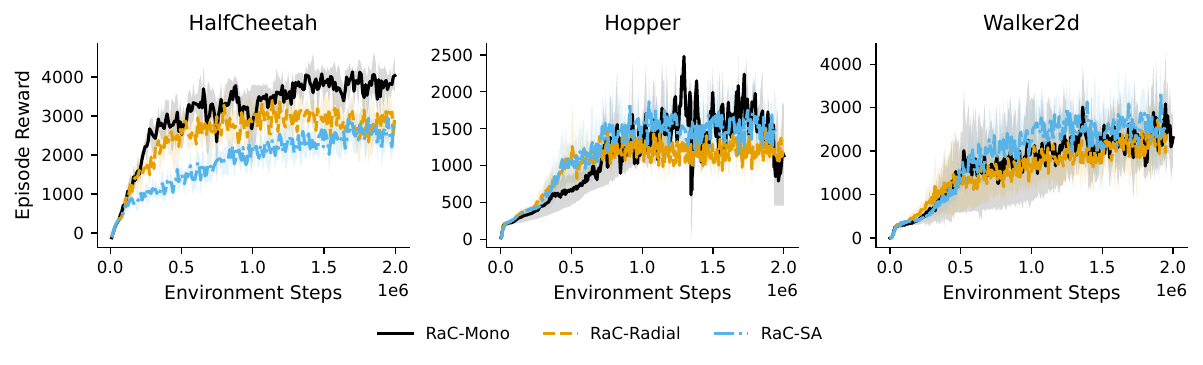}
    \caption{Training curves of the RaC methods. }
    \Description{Training curves of the RaC methods. }
    \label{fig:convergence}
\end{figure*}
The constraint surrogates derived above contain terms involving the divergence between the next policy and the current policy, i.e., $D_{\mathrm{TV}}(\pi\Vert\pi_k)$. In practice, we omit these terms and control the divergence between successive policies through the policy-divergence constraint already included in Eq.~\eqref{eq:constrained_robustness}, i.e., $D(\pi,\pi_k) \leq \delta$. This follows the standard practice in trust-region methods in which policy divergence is controlled using a separate constraint rather than through divergence penalties~\cite{DBLP:conf/icml/AchiamHTA17,DBLP:conf/icml/SchulmanLAJM15}.

Moreover, keeping both the standard policy-divergence terms and the worst-case divergence terms could result in overly conservative updates. We retain the worst-case terms as they directly capture the robustness property targeted by our constraints. This relaxation improves training stability at the cost of potential constraint violations, which we characterize in Proposition \ref{prop-worst-case-violation}.

\begin{proposition}
\label{prop-worst-case-violation}
\textbf{(Worst-case constraint violation)}
Suppose that $\pi_k$ and $\pi_{k+1}$ are related by Eq.~\eqref{eq:constrained_robustness}, and that the standard policy-divergence terms are omitted from the constraint surrogates, with the divergence between successive policies controlled only through $D(\pi_{k+1},\pi_k) \leq \delta$, and that
$\pi_{k+1}$ satisfies the resulting surrogate constraint. Then, the following worst-case constraint-violation bounds hold:

\begin{itemize}
    \item For \textit{RaC-Mono}:
    \begin{equation}
        J(\pi_{k+1}^{\mathrm{wc}})
        - J(\pi_k^{\mathrm{wc}})
        \geq
        -\gamma\alpha_{\mathrm{wc}}^A
        \E_{s\sim d^{\pi_k}}
        \left[
            D_{\mathrm{TV}}(\pi_{k+1}\Vert\pi_k)[s]
        \right]
    \end{equation}

    \item For \textit{RaC-SA}:
    \begin{equation}
        J(\pi_{k+1})
        - J(\pi_{k+1}^{\mathrm{wc}})
        \leq
        \Delta_{\mathrm{max}}
        + 2\alpha^R
        \E_{s\sim d^{\pi_k}}
        \left[
            D_{\mathrm{TV}}(\pi_{k+1}\Vert\pi_k)[s]
        \right]
    \end{equation}

    \item For \textit{RaC-Radial}:
    \begin{equation}
        J(\pi_{k+1})
        - J(\pi_{k+1}^{\mathrm{wc}})
        \leq
        \Delta_{\mathrm{max}}
        + \alpha^{AR}
        \E_{s\sim d^{\pi_k}}
        \left[
            D_{\mathrm{TV}}(\pi_{k+1}\Vert\pi_k)[s]
        \right]
    \end{equation}
\end{itemize}
\end{proposition}

To solve these constrained problems, we follow PPO and replace the explicit policy-divergence constraint $D(\pi,\pi_k)\leq\delta$ with a clipped surrogate objective~\cite{DBLP:journals/corr/SchulmanWDRK17}. We choose PPO for its computational efficiency, since both our methods and the regularization-based methods reviewed above require convex relaxation computations. Its first-order updates avoid the second-order computations and line search used by TRPO~\cite{DBLP:conf/icml/SchulmanLAJM15} or CPO~\cite{DBLP:conf/icml/AchiamHTA17}. We use PPO-style clipping in the nominal objective and, for \emph{RaC-Mono} and \emph{RaC-Radial}, in the advantage-based robustness surrogates.

We next explain how the retained worst-case divergence terms are handled during training. Prior work generally avoids using constants such as $\alpha^R$ or $\alpha^A$ directly in the training objective, as they can be large and therefore place excessive weight on constraint satisfaction at the expense of policy optimization. In contrast, we estimate these coefficients and use them in our implementation, with the following modifications:
\begin{enumerate}
    \item  We approximate $\max_{s,a,s'}|R(s,a,s')|$ and $\max_{s,a}|A(s,a)|$ with the maximum absolute values observed in the trajectories collected under the current policy $\pi_k$, rather than over the entire state-action space.
    \item We balance the return and constraint gradients following~\cite{DBLP:conf/icml/StookeAA20}. For example, for RaC-SA, the policy update uses the direction
    \begin{equation}
        \nabla_\theta J
        -
        \lambda\beta\nabla_\theta\mathcal{C}
        \label{eq:gradient_balance}
    \end{equation}
    where
    \[
        \beta =
        \frac{\|\nabla_\theta J\|}
        {\|\nabla_\theta\mathcal{C}\|}
    \]
    This makes the return and scaled constraint gradients equal in magnitude before weighting by $\lambda$, so that $\lambda$ controls their relative contribution. The scaling is applied only to the policy update. The policy parameters are updated using minibatches, while $\lambda$ is kept fixed during these updates. The multiplier $\lambda$ is not updated on individual minibatches; it is updated separately using the unscaled constraint value estimated from all trajectories collected under $\pi_k$.
\end{enumerate}

With these modifications, the algorithms trained stably and achieved strong robustness in our experiments. Finally, using Pinsker inequality, $\E[D_{\mathrm{TV}}] \leq \sqrt{\frac{1}{2}\E[D_{\mathrm{KL}}]}$, we replace the total variation divergence with its KL-based upper bound.

\section{Experiments}

\subsection{Experimental Setup}

We complement our theoretical analysis with experiments on the MuJoCo benchmark~\cite{DBLP:conf/nips/TowersKBCDGKKKP25}. Our evaluation examines the robustness of combined regularizations and the proposed constrained methods. For a fair comparison, all algorithms are trained for the same number of environment steps (2 million) and use an identical policy network: a two-layer MLP with 64 hidden units per layer. We perform a grid search to select the best hyperparameters.

To evaluate robustness, we use three state-of-the-art adaptive attacks: Robust SARSA (RS)~\cite{DBLP:conf/nips/0001CX0LBH20}, SA-RL~\cite{DBLP:conf/iclr/ZhangCBH21}, and PA-AD~\cite{DBLP:conf/iclr/SunZLH22}. These are \emph{learned attacks}, meaning that they must first be trained before being used to generate adversarial perturbations. RS uses TD learning to train an action-value network for the victim agent, whereas SA-RL and PA-AD train a policy using deep RL to generate perturbations. We use PPO to train both SA-RL and PA-AD. We use the standard environment-specific perturbation budgets adopted in prior work~\cite{DBLP:conf/nips/0001CX0LBH20,DBLP:conf/nips/OikarinenZMDW21,DBLP:conf/nips/LiangSZH22} to maintain comparability with established robustness evaluations. 

These attacks are sensitive to the choice of hyperparameters. As a result, we evaluate against a large and diverse population of attackers. For each method and environment, we evaluate against 2328 attackers: 1296 PA-AD attackers, 600 SA-RL attackers, and 432 RS attackers. Each attacker corresponds to a different hyperparameter configuration. Prior work evaluates robustness against significantly fewer adversaries (e.g., 30, 50, or 216). Each attack is evaluated over 1000 episodes (20 seeds, 50 episodes per seed). We report the interquartile mean (IQM)~\cite{DBLP:conf/nips/AgarwalSCCB21} and the standard deviation of the episodic returns.

All experiments are conducted on Intel Xeon Platinum CPUs. The adversarial evaluation requires substantial computation, as the learned attackers are trained separately for each victim policy. A complete attacker sweep for one method in one environment requires approximately 935 CPU-hours for PA-AD, 275 CPU-hours for SA-RL, and 172 CPU-hours for RS, totaling 1382 CPU-hours. Evaluating all eight methods in a single environment therefore requires approximately 11056 CPU-hours. Across the three environments, this amounts to approximately 33168 CPU-hours. These numbers cover only the adversarial evaluation of the agents reported in Table~\ref{tab:robustness-full}. They exclude the policy-training hyperparameter sweeps and the additional adversarial evaluations used to select the reported agents. We provide further experimental details in the appendix.

Table~\ref{tab:robustness-full} reports the nominal and attacked returns of the evaluated methods. For each method and environment, we first select agents from the hyperparameter sweep based on their nominal performance. We then test them against 40 RS attackers and discard those that perform poorly. We run the full adversarial evaluation on the remaining agents and report the one with the best performance under attack. Within each environment, we group each base method with its SA-Reg combination, where applicable, and its corresponding RaC variant. The \emph{Worst across attacks} column reports the minimum return across the three attack families, and \emph{Gain vs. base} reports its improvement relative to the base method.
\begin{table*}[t]
    \centering
    \caption{Nominal and adversarial performance.}
    \setlength{\tabcolsep}{4pt}
    \begin{tabular}{llcccccc}
    \toprule
    \multirow{2}{*}{Env} & \multirow{2}{*}{Method} & \multirow{2}{*}{\shortstack{Nominal\\Performance}} & \multicolumn{3}{c}{Attack} & \multirow{2}{*}{\shortstack{Worst across\\attacks}} & \multirow{2}{*}{\shortstack{Gain vs. \\base}}\\
    \cmidrule(lr){4-6}
     & & & RS & SA-RL & PA-AD & \\
    \midrule
    \multirow{8}{*}{\shortstack{HalfCheetah\\$(\epsilon = 0.15)$}}
        & WocaR-RL      & $5666.4 \pm 412$ & $401.4 \pm 252$ & $-88.8 \pm 162$ & $50.8 \pm 903$ & $-88.8$ & -- \\
        & WocaR-RL + SA      & $4786.4 \pm 97$ & $4396.7 \pm 67$ & $4469.38 \pm 156$ & $4458.2 \pm 205$ & $4396.7$& \gain{+4485.5} \\
        & RaC-Mono   & $5334.7 \pm 68$ & $4670.1 \pm 204$ & $4697.7 \pm 174$ & $  \mathbf{4871.7 \pm 384}$ & $4670.1$ & \gain{+4758.9} \\
        \cmidrule(lr){2-8}
        & Radial  & $4053.8 \pm 522$ & $4028.1 \pm 533$ & $4037.2 \pm 472$ & $4033.1 \pm 536$ & $4028.1$ & -- \\
        & Radial + SA  & $5272 \pm 143$ & $4613.6 \pm 130$ & $4620.3 \pm 137$ & $4780 \pm 475$ & $4613.6$ & \gain{+585.5}\\
        & RaC-Radial & $5902.8 \pm 54$ & $4680.6 \pm 74$ & $3727.8 \pm 1195$ & $4510 \pm 1511$ & $3727.8$ & \loss{-300.3}\\
        \cmidrule(lr){2-8}
        & SA         & $4903.3 \pm 36$ & $3795.7 \pm 92$ & $3940.0 \pm 601$ & $3796.9 \pm 41$ & $3795.7$ & --\\
        & RaC-SA     & $4983.1 \pm 76$ & $\mathbf{4801.8 \pm 93}$ & $\mathbf{4818.3 \pm 82}$ & $4831.2 \pm 88$ & $\mathbf{4801.8}$ & \gain{+1006.1} \\
    \cmidrule(lr){1-8}
    \multirow{8}{*}{\shortstack{Hopper\\$(\epsilon = 0.075)$}}
        & WocaR-RL      & $3317.3 \pm 1$ & $595.9 \pm 499$ & $761.4 \pm 60$ & $727.3 \pm 150$ & $595.9$ & --\\
        & WocaR-RL + SA      & $3146.6 \pm 712$ & $1150.7 \pm 48$ & $1279.9 \pm 97$ & $1217.7 \pm 196$ & $1150.7$ & \gain{+554.8}\\
        & RaC-Mono   & $2782 \pm 36$ & $1156.8 \pm 190$ & $1216.7 \pm 200$ & $1366.5 \pm 299$ & $1156.8$ & \gain{+ 560.9}\\
        \cmidrule(lr){2-8}
        & Radial  & $3487.2 \pm 76$ & $858.6 \pm 52$ & $888.07 \pm 118$ & $937 \pm 162$ & $858.6$ & -- \\
        & Radial + SA  & $2877 \pm 537$ & $1056.6 \pm 964$ & $1053.6 \pm 864$ & $1054.3 \pm 940$ & $1053.6$  & \gain{+ 195}\\
        & RaC-Radial & $2617.1 \pm 692$ & $\mathbf{1525.8 \pm 266}$ & $\mathbf{1487.1 \pm 314}$ & $\mathbf{1525 \pm 254}$ & $\mathbf{1487.1}$ & \gain{+ 628.5}\\
        \cmidrule(lr){2-8}
        & SA         & $3555.2 \pm 5$ & $1363.32 \pm 265$ & $1128.5 \pm 917$ & $1140.5 \pm 656$ & $1128.5$ & --\\
        & RaC-SA     & $2275.6 \pm 857$ & $1226.5 \pm 918$ & $1073.8 \pm 717$ & $1111.6 \pm 644$ & $1073.8$ & \loss{- 54.7} \\
    \cmidrule(lr){1-8}
    \multirow{8}{*}{\shortstack{Walker2d\\$(\epsilon = 0.05)$}}
        & WocaR-RL      & $6402.7 \pm 294$ & $184.9 \pm 1907$ & $257 \pm 1603$ & $284.3 \pm 2294$ & $184.9$ & --\\
        & WocaR-RL + SA & $4799.8 \pm 161$ & $4444.1 \pm 842$ & $4470 \pm 181$ & $4465.9 \pm 149$ & $4444.1$ & \gain{+4259.2} \\
        & RaC-Mono   & $4433.8 \pm  678$ & $2666.8 \pm 1501$ & $3108.2 \pm 1354$ & $1603.3 \pm 1336$ & $1603.3$ & \gain{+1418.4} \\
        \cmidrule(lr){2-8}
        & Radial & $3114 \pm 188$  & $543.3 \pm 658$ & $792.3 \pm 408$ & $824.4 \pm 425$ & $543.3$ & -- \\
        & Radial + SA  & $4588 \pm 710$ & $4133.6 \pm 851$ & $4117.6 \pm 561$ & $3977.2 \pm 1357$ & $3977.2$ & \gain{+3433.9} \\
        & RaC-Radial & $5585.3 \pm 497$ & $4849 \pm 2002$ & $\mathbf{4845.1 \pm 1791}$ & $\mathbf{4644 \pm 1333}$ & $\mathbf{4644}$ & \gain{+4100.7}\\
        \cmidrule(lr){2-8}
        & SA         & $6129 \pm 883$ & $802.6 \pm 1691$ & $400.9 \pm 2097$ & $591 \pm 2123$ & $400.9$ & --\\
        & RaC-SA     & $5763.1 \pm 582$ & $\mathbf{4858.4 \pm 1823}$ & $3865.9 \pm 1335$ & $4338.5 \pm 1614$ & $3865.9$ & \gain{+3465} \\
    \bottomrule
    \end{tabular}
    \label{tab:robustness-full}
\end{table*}

\subsection{Combined Regularization}

We begin by providing empirical evidence that SA-Reg improves the robustness of both Radial and WocaR-RL. Across all environments and attack families, adding SA-Reg improves the attacked returns of both methods. In particular, WocaR-RL did not perform well under attack when used alone. After combining it with SA-Reg, we observe large gains in robustness. One of the main contributions of our work is to provide a theoretical justification for combining these robustness methods.

\subsection{Constrained Robustness}
In this section, we provide an empirical analysis of the RaC algorithms: RaC-Mono, RaC-SA, and RaC-Radial. Specifically, we evaluate their convergence, robustness, and computational cost.

\subsubsection{Convergence} Figure~\ref{fig:convergence} shows the learning curves of the three methods. All three methods converge during training across all environments. Note that this figure reports the episodic returns of the stochastic policies. The episodic returns of the final deterministic policies are reported in Table~\ref{tab:robustness-full} under the \emph{Nominal Performance} column. Notably, the training of RaC methods does not collapse, despite explicitly using coefficients such as $\alpha^{R}$ or $\alpha^{A}$.

\subsubsection{Robustness}
We present the results of the adversarial evaluation in Table~\ref{tab:robustness-full}. Within each environment, we highlight in bold the highest return against each attack family and the highest minimum return across attack families, reported in the \emph{Worst across attacks} column. The main takeaways are as follows:

\begin{enumerate}
    \item Overall, the RaC family achieves the best robust performance in every environment: \emph{RaC-SA} in HalfCheetah and \emph{RaC-Radial} in Hopper and Walker2d. A RaC method also achieves the highest return in every individual attack column. These results show that the proposed constrained methods can achieve strong robustness across different environments and attack families.
    
    \item RaC-SA outperforms SA in every environment under every attack except in Hopper where their performance is very close. Its minimum return across attacks improves by $1006.1$ and $3465.0$ in HalfCheetah and Walker2d, respectively. Similarly, RaC-Radial outperforms Radial + SA except for two attacks in the HalfCheetah environment. These comparisons provide empirical evidence for the effectiveness of using gap tolerance as a constraint for robustness.
    
    \item RaC-Mono outperforms WocaR-RL under every attack in all three environments. Compared with WocaR-RL + SA, however, its results are mixed: it achieves higher returns under all three attacks in HalfCheetah, comparable performance in Hopper, and lower returns in Walker2d. The results suggest that the objective of monotonically improving worst-case performance alone may not provide a sufficiently strong or appropriate inductive bias for learning robust policies.
\end{enumerate}

\subsubsection{Sensitivity to tolerance $\eta$}
We study the sensitivity of RaC-SA and RaC-Radial to the tolerance parameter $\eta$. Table~\ref{tab:tolerance} reports the results under the RS attack when the maximum tolerated performance degradation is set to $10\%$, $20\%$, or $50\%$. Among the tested values, $\eta = 20\%$ provides the most consistent performance across both methods and all environments. A tighter tolerance does not necessarily improve robustness: at $\eta = 10\%$, RaC-Radial performs poorly in HalfCheetah. Conversely, at $\eta = 50\%$, RaC-SA exhibits substantial performance drops in Hopper and Walker2d.

\subsubsection{Wall-clock time}
\balance
Table~\ref{tab:wallclock} reports the wall-clock training time of the RaC algorithms alongside their corresponding baselines. RaC-SA and RaC-Radial require approximately 19 and 29 minutes more training time than SA-Reg and Radial + SA-Reg, respectively. This additional overhead is due to the computation of the gradient balancing coefficient $\beta$ used in Eq.~\eqref{eq:gradient_balance}. Given the robustness improvements of RaC-SA and RaC-Radial shown in Table~\ref{tab:robustness-full}, we believe that this additional computational cost is justified.

\begin{table}[t]
    \centering
    \caption{Sensitivity to tolerance $\eta$.}
    \setlength{\tabcolsep}{1mm}
    \begin{tabular}{lccc}
    \toprule
    $\eta$ & $10\%$ & $20\%$ & $50\%$ \\
    \midrule
    \multicolumn{4}{l}{\textit{HalfCheetah}} \\
    RaC-SA & $4476 \pm 64$ & $4435 \pm 125$ & $\mathbf{4802} \pm 93$ \\
    RaC-Radial  & $888 \pm 1086$  & $\mathbf{4680} \pm 74$  & $3769 \pm 1076$ \\
    \cmidrule(lr){1-4}
    \multicolumn{4}{l}{\textit{Hopper}} \\
    RaC-SA & $1057 \pm 10$ & $\mathbf{1226} \pm 918$ & $610 \pm 1170$ \\
    RaC-Radial  & $1396 \pm 106$ & $\mathbf{1525.8} \pm 266$ & $1151 \pm 230$ \\
    \cmidrule(lr){1-4}
    \multicolumn{4}{l}{\textit{Walker2d}} \\
    RaC-SA & $\mathbf{4858} \pm 1823$ & $4564 \pm 165$ & $1679 \pm 1520$\\
    RaC-Radial  & $\mathbf{4849} \pm 2002$ & $4783 \pm 1164$ & $3599 \pm 1409$ \\
    \bottomrule
    \end{tabular}
    \label{tab:tolerance}
\end{table}

\begin{table}[tb]
    \centering
    \caption{Training wall-clock time for each method.}
    \begin{tabular}{lc}
    \toprule
    Method & Time (h) \\
    \midrule
    WocaR-RL      & 2.49 $\pm$ 0.09 \\
    WocaR-RL + SA      & 2.76 $\pm$ 0.1 \\
    RaC-Mono   & 3.21 $\pm$ 0.06 \\
    Radial     & 2.4 $\pm$ 0.01 \\ 
    Radial + SA  & 2.53 $\pm$ 0.01 \\
    RaC-Radial & 3.01 $\pm$ 0.02 \\
    SA         & 2.19 $\pm$ 0.02 \\
    RaC-SA     & 2.5 $\pm$ 0.01\\
    \bottomrule
    \end{tabular}
    \label{tab:wallclock}
    \end{table}

\section{Conclusion}
In this paper, we make two theoretical contributions to adversarial reinforcement learning. First, we unify existing regularization techniques by deriving new upper bounds on the gap between nominal and worst-case performance. Furthermore, the SA-Reg penalty appears in the new bounds, providing a theoretical justification for combining Radial or WocaR-RL with SA-Reg. Second, we formulate robust training as a constrained policy optimization problem. Building on our new bounds, we derive tractable surrogates for the robustness constraints and treat the choice of the regularization weight $\lambda$ as part of the learning problem, updating it jointly with the policy rather than keeping it fixed throughout training. We use this framework to develop three new algorithms: RaC-Mono, RaC-SA, and RaC-Radial. Finally, we complement our theoretical analysis with a thorough adversarial evaluation using a large and diverse population of adaptive attackers. The results demonstrate the robustness gains from combining regularizations, while RaC-SA and RaC-Radial achieve the best robust performance across the evaluated environments.





\bibliographystyle{ACM-Reference-Format} 
\bibliography{sample}


\newpage


\section*{Supplementary material}

# Technical Appendix

## A Background

**Definition 1.** *(Worst-case RL functions, from Liang et al. (2022))*
*We define the worst-case value of $\pi$ as :*

$$V^{\pi}_{\mathrm{wc}}(s) = \mathop{\mathbb{E}}_{\tau\sim\pi}\left[\sum_{t=0}^{\infty}\gamma^t R(s_t, a_t \sim \pi(.|h^*(s_t)), s_{t+1})|s_0 = s\right] \tag{1}$$

*Similarly, we define the worst-case action-value function:*

$$Q^{\pi}_{\mathrm{wc}}(s,a) = \mathop{\mathbb{E}}_{\tau\sim\pi}\left[\sum_{t=0}^{\infty}\gamma^t R(s_t, a_t \sim \pi(.|h^*(s_t)), s_{t+1})|s_0 = s, a_0 = a\right] \tag{2}$$

*Finally, we define the worst-case advantage function as:* $A^{\pi}_{\mathrm{wc}}(s,a) = Q^{\pi}_{\mathrm{wc}}(s,a) - V^{\pi}_{\mathrm{wc}}(s)$

**Definition 2.** *(Worst-case Bellman Equation, from (Zhang et al. (2020)))*
*The worst-case RL functions obey the following Bellman equations:*

$$V^{\pi}_{\mathrm{wc}}(s) = \sum_{a\in\mathcal{A}}\pi^{\mathrm{wc}}(a|s)\sum_{s'\in\mathcal{S}}P(s'|s,a)\left(R(s,a,s') + \gamma V^{\pi}_{\mathrm{wc}}(s')\right) \tag{3}$$

$$Q^{\pi}_{\mathrm{wc}}(s,a) = \sum_{s'\in\mathcal{S}}P(s'|s,a)\left(R(s,a,s') + \gamma\sum_{a'\in\mathcal{A}}\pi^{\mathrm{wc}}(a'|s')Q^{\pi}_{\mathrm{wc}}(s',a')\right) \tag{4}$$

**Definition 3.** *(Worst-case Bellman Optimality Equation, from (Liang et al. (2022)) )*
*The worst-case action-value function obeys the following Bellman Optimality equation :*

$$Q^{\pi}_{\mathrm{wc}}(s,a) = \mathop{\mathbb{E}}_{s'\sim P}\left[R(s,a,s') + \gamma\min_{a'\in\mathcal{A}_{\mathrm{adv}}(s',\pi)}Q^{\pi}_{\mathrm{wc}}(s',a')\right] \tag{5}$$

*The worst-case value function can be obtained using:*

$$V^{\pi}_{\mathrm{wc}}(s) = \min_{a\in\mathcal{A}_{\mathrm{adv}}(s,\pi)}Q^{\pi}_{\mathrm{wc}}(s,a) \tag{6}$$

*Here $\mathcal{A}_{\mathrm{adv}} \doteq \{a \in \mathcal{A} : \exists\tilde{s} \in \mathcal{B}_\epsilon(s) \text{ s.t } \pi(\tilde{s}) = a\}$ is the set the admissible of actions an adversary can mislead the victim into selecting with the $\epsilon$ budget.*

**Theorem 1.** *Achiam et al. (2017)*
*For any function $f : S \to \mathbb{R}$ and any policies $\pi'$ and $\pi$, define $\delta_f(s,a,s') \doteq R(s,a,s') + \gamma f(s') - f(s)$, and $\epsilon_f^{\pi'} \doteq \max_s |\mathbb{E}_{a\sim\pi', s'\sim P}[\delta_f(s,a,s')]|$, and:*

$$L_{\pi,f}(\pi') \doteq \mathbb{E}_{\substack{s\sim d^\pi \\ a\sim\pi \\ s'\sim P}}\left[\left(\frac{\pi'(a|s)}{\pi(a|s)} - 1\right)\delta_f(s,a,s')\right],$$

$$D^{\pm}_{\pi,f}(\pi') \doteq \frac{L_{\pi,f}(\pi')}{1-\gamma} \pm \frac{2\gamma\epsilon_f^{\pi'}}{(1-\gamma)^2}\mathbb{E}_{s\sim d^\pi}\left[D_{TV}(\pi\|\pi')[s]\right].$$

*Here $D_{TV}$ is the total variation divergence. The following bounds hold*

$$D^{-}_{\pi,f}(\pi') \leq J(\pi') - J(\pi) \leq D^{+}_{\pi,f}(\pi') \tag{7}$$

Throughout this Appendix, we use the following two lemmas:

**Lemma 1.** *Given the worst-case policy and the worst-case advantage, the following holds:*

$$\sum_{a\in\mathcal{A}} \pi^{\text{wc}}(a|s)A^{\pi}_{\text{wc}}(s,a) = 0 \tag{8}$$

*Proof.* The lemma is a direct consequence of Definition 1 and Bellman equations in Definition 2. □

**Lemma 2** (Triangle inequality of the total variation distance)**.** *Let $\mu$, $\nu$, and $\rho$ be three probability distributions, then:*

$$D_{TV}(\mu,\rho) \leq D_{TV}(\mu,\nu) + D_{TV}(\nu,\rho) \tag{9}$$

*Proof.* By the definition of the total variation distance:

$$D_{TV}(\mu,\rho) = \frac{1}{2}\sum_{x} |\mu(x) - \rho(x)|$$

Using the triangle inequality of the absolute value:

$$|\mu(x) - \rho(x)| \leq |\mu(x) - \nu(x)| + |\nu(x) - \rho(x)|$$

Summing over all $x$ yields:

$$D_{TV}(\mu,\rho) \leq D_{TV}(\mu,\nu) + D_{TV}(\nu,\rho),$$

□

# B Theoretical Proofs

## B.1 Proof of Theorem 2:SA-Reg

In this subsection, we provide the proof for the upper bound used by SA-Reg (Zhang et al. (2020)). We use Theorem 1, and select the following function $\forall s \in S : f(s) = 0$, and $\pi' = \pi^{\text{wc}}$. Therefore we have:

$$J(\pi) - J(\pi^{\text{wc}}) \leq \frac{-1}{1-\gamma} L_{\pi,0}(\pi^{\text{wc}}) + \frac{2\gamma\epsilon_0^{\pi^{\text{wc}}}}{(1-\gamma)^2}\mathbb{E}_{s\sim d^{\pi}}[D_{TV}(\pi\|\pi^{\text{wc}})[s]]$$

For the first term:

$$\begin{aligned}
\frac{-1}{1-\gamma} L_{\pi,0}(\pi^{\text{wc}}) &= \frac{-1}{1-\gamma}\mathbb{E}_{\substack{s\sim d^{\pi}\\ a\sim\pi\\ s'\sim P}}\left[\left(\frac{\pi^{\text{wc}}(a|s)}{\pi(a|s)} - 1\right)\delta_0(s,a,s')\right] \\
&= \frac{-1}{1-\gamma}\sum_{s} d^{\pi}(s)\sum_{a}\pi(a|s)\sum_{s'}P(s'|s,a)\left[\left(\frac{\pi^{\text{wc}}(a|s)}{\pi(a|s)} - 1\right)R(s,a,s')\right] \\
&= \frac{1}{1-\gamma}\sum_{s} d^{\pi}(s)\sum_{a}(\pi(a|s) - \pi^{\text{wc}}(a|s))\sum_{s'}P(s'|s,a)R(s,a,s') \\
&\leq \frac{1}{1-\gamma}\sum_{s} d^{\pi}(s)\sum_{a}|\pi(a|s) - \pi^{\text{wc}}(a|s)|\sum_{s'}P(s'|s,a)\,|R(s,a,s')| \\
&\leq \frac{1}{1-\gamma}\sum_{s} d^{\pi}(s)\sum_{a}|\pi(a|s) - \pi^{\text{wc}}(a|s)|\sum_{s'}P(s'|s,a)\max_{s,a,s'}|R(s,a,s')| \\
&\leq \frac{\max_{s,a,s'}|R(s,a,s')|}{1-\gamma}\sum_{s} d^{\pi}(s)\sum_{a}|\pi(a|s) - \pi^{\text{wc}}(a|s)| \\
&\leq \frac{2\max_{s,a,s'}|R(s,a,s')|}{1-\gamma}\sum_{s} d^{\pi}(s)D_{TV}(\pi\|\pi^{\text{wc}})[s] \\
&\leq \frac{2\max_{s,a,s'}|R(s,a,s')|}{1-\gamma}\mathbb{E}_{s\sim d^{\pi}}[D_{TV}(\pi\|\pi^{\text{wc}})[s]]
\end{aligned}$$

For the second term:

$$\frac{2\gamma\epsilon_0^{\pi^{\text{wc}}}}{(1-\gamma)^2}\mathbb{E}_{s\sim d^\pi}[D_{TV}(\pi\|\pi^{\text{wc}})[s]] = \frac{2\gamma\max_s|\mathbb{E}_{a\sim\pi^{\text{wc}},\, s'\sim P}[R(s,a,s')]|}{(1-\gamma)^2}\mathbb{E}_{s\sim d^\pi}[D_{TV}(\pi\|\pi^{\text{wc}})[s]]$$
$$\leq \frac{2\gamma\max_{s,a,s'}|R(s,a,s')|}{(1-\gamma)^2}\mathbb{E}_{s\sim d^\pi}[D_{TV}(\pi\|\pi^{\text{wc}})[s]]$$

Therefore

$$J(\pi)-J(\pi^{\text{wc}}) \leq \frac{2\max_{s,a,s'}|R(s,a,s')|}{1-\gamma}\mathbb{E}_{s\sim d^\pi}[D_{TV}(\pi\|\pi^{\text{wc}})[s]] + \frac{2\gamma\max_{s,a,s'}|R(s,a,s')|}{(1-\gamma)^2}\mathbb{E}_{s\sim d^\pi}[D_{TV}(\pi\|\pi^{\text{wc}})[s]]$$
$$\leq 2\max_{s,a,s'}|R(s,a,s')|\left(\frac{1}{1-\gamma}+\frac{\gamma}{(1-\gamma)^2}\right)\mathbb{E}_{s\sim d^\pi}[D_{TV}(\pi\|\pi^{\text{wc}})[s]]$$
$$\leq \frac{2\max_{s,a,s'}|R(s,a,s')|}{(1-\gamma)^2}\mathbb{E}_{s\sim d^\pi}[D_{TV}(\pi\|\pi^{\text{wc}})[s]]$$

Hence, we end up with the following upper bound:

**SA-Reg Bound**

$$J(\pi)-J(\pi^{\text{wc}}) \leq \frac{2\max_{s,a,s'}|R(s,a,s')|}{(1-\gamma)^2}\mathbb{E}_{s\sim d^\pi}[D_{TV}(\pi\|\pi^{\text{wc}})[s]] \tag{10}$$

There are two remarks regarding the bound shown in Eq. (10) and the bound provided in the original paper Zhang et al. (2020). First, the original paper contains a minor error. Instead of having

$$\frac{2\max_{s,a,s'}|R(s,a,s')|}{(1-\gamma)^2}$$

the bound is given as

$$2\left(1+\frac{\gamma}{(1-\gamma)^2}\right)\max_{s,a,s'}|R(s,a,s')|$$

due to the omission of the factor $\frac{1}{1-\gamma}$ in the first term above. Second, we retain the expectation of $D_{TV}$ rather than its maximum over the state space, as it is the expectation that is used in the implementation.

Given that we only have samples collected with the current policy $\pi_k$, the above upper bound cannot be easily estimated as it depends on $d^\pi$ and not $d^{\pi_k}$. Therefore, we have an interest in having an upper bound that depends on $d^{\pi_k}$. Below, we provide the proof for such an upper bound.

First, by the Eq. (10) and by replacing $\pi$ with $\pi_k$ we have:

$$J(\pi_k)-J(\pi^{\text{wc}}) \leq \frac{2\max_{s,a,s'}|R(s,a,s')|}{(1-\gamma)^2}\mathbb{E}_{s\sim d^{\pi_k}}[D_{TV}(\pi_k\|\pi^{\text{wc}})[s]] \tag{11}$$

Moreover, if we use Theorem 1 and start by $J(\pi')-J(\pi)\leq D^+_{\pi,f}(\pi')$ and use the same procedure above, we will find the following inequality:

$$J(\pi')-J(\pi) \leq \frac{2\max_{s,a,s'}|R(s,a,s')|}{(1-\gamma)^2}\mathbb{E}_{s\sim d^\pi}[D_{TV}(\pi\|\pi')[s]]$$

Therefore, we have (replace $\pi'$ by $\pi$ and $\pi$ by $\pi_k$):

$$J(\pi)-J(\pi_k) \leq \frac{2\max_{s,a,s'}|R(s,a,s')|}{(1-\gamma)^2}\mathbb{E}_{s\sim d^{\pi_k}}[D_{TV}(\pi_k\|\pi)[s]] \tag{12}$$

Combining Eq.(11) and Eq.(12), we find:

$$J(\pi)-J(\pi^{\text{wc}}) \leq \alpha^R\mathbb{E}_{s\sim d_k^\pi}[D_{TV}(\pi_k\|\pi)[s]] + \alpha^R\mathbb{E}_{s\sim d_k^\pi}[D_{TV}(\pi_k\|\pi^{\text{wc}})[s]] \tag{13}$$

Here $\alpha^R = \frac{2\max_{s,a,s'}\left|R(s,a,s')\right|}{(1-\gamma)^2}$.

In the RHS of Eq.( 13) we have $D_{TV}(\pi_k\|\pi^{\mathrm{wc}})$. However, the existing implementations use $D_{TV}(\pi\|\pi^{\mathrm{wc}})$ (Experiments also show that using $D_{TV}(\pi_k\|\pi^{\mathrm{wc}})$ is less effective). Therefore, we use the triangular inequality of $D_{TV}$ to get the following:

$$J(\pi) - J(\pi^{\mathrm{wc}}) \leq 2\alpha^R \mathbb{E}_{s\sim d_k^\pi}[D_{TV}(\pi_k\|\pi)[s]] + \alpha^R \mathbb{E}_{s\sim d_k^\pi}[D_{TV}(\pi\|\pi^{\mathrm{wc}})[s]] \tag{14}$$

To compute $D_{TV}(\pi\|\pi^{\mathrm{wc}})[s]$, SA-Reg uses:

$$D_{TV}(\pi\|\pi^{\mathrm{wc}})[s] = \max_{\tilde{s}\in B(s)} D_{TV}(\pi(.|s)\|\pi(.|\tilde{s})) \tag{15}$$

## B.2 Proof of Theorem 3: Radial Inspired Bound

In this subsection, we provide a new bound inspired by the one used in Radial (Oikarinen et al. (2021)). Our new bound also justifies why SA-Reg is helpful for Radial.

We use Theorem 1, with the following function $\forall s \in S : f(s) = V^\pi(s)$ and $\pi' = \pi^{\mathrm{wc}}$. We have :

$$J(\pi) - J(\pi^{\mathrm{wc}}) \leq \frac{-1}{1-\gamma} L_{\pi,V^\pi}(\pi^{\mathrm{wc}}) + \frac{2\gamma\epsilon_{V^\pi}^{\pi^{\mathrm{wc}}}}{(1-\gamma)^2}\mathbb{E}_{s\sim d^\pi}[D_{TV}(\pi\|\pi^{\mathrm{wc}})[s]]$$

For the first term:

$$\begin{aligned}
\frac{-1}{1-\gamma}L_{\pi,V^\pi}(\pi^{\mathrm{wc}}) &= \frac{-1}{1-\gamma}\mathbb{E}_{\substack{s\sim d^\pi\\ a\sim\pi\\ s'\sim P}}\left[\left(\frac{\pi^{\mathrm{wc}}(a|s)}{\pi(a|s)} - 1\right)\delta_{V^\pi}(s,a,s')\right] \\
&= \frac{-1}{1-\gamma}\sum_s d^\pi(s)\sum_a \pi(a|s)\sum_{s'} P(s'|s,a)\left[\left(\frac{\pi^{\mathrm{wc}}(a|s)}{\pi(a|s)} - 1\right)\delta_{V^\pi}(s,a,s')\right] \\
&= \frac{1}{1-\gamma}\sum_s d^\pi(s)\sum_a (\pi(a|s) - \pi^{\mathrm{wc}}(a|s))\underbrace{\sum_{s'} P(s'|s,a)\left(R(s,a,s') + \gamma V^\pi(s') - V^\pi(s)\right)}_{=A^\pi(s,a)} \\
&= \frac{1}{1-\gamma}\sum_s d^\pi(s)\sum_a (\pi(a|s) - \pi^{\mathrm{wc}}(a|s))\,A^\pi(s,a) \\
&= \frac{1}{1-\gamma}\sum_s d^\pi(s)\sum_a (\pi(a|s) - \pi^{\mathrm{wc}}(a|s))\,A^\pi(s,a) \\
&= \frac{1}{1-\gamma}\sum_s d^\pi(s)\underbrace{\sum_a \pi(a|s)A^\pi(s,a)}_{=0} - \frac{1}{1-\gamma}\sum_s d^\pi(s)\sum_a \pi^{\mathrm{wc}}(a|s)A^\pi(s,a) \\
&= \frac{-1}{1-\gamma}\sum_s d^\pi(s)\sum_a \pi^{\mathrm{wc}}(a|s)A^\pi(s,a) \\
&= \frac{-1}{1-\gamma}\mathop{\mathbb{E}}_{\substack{s\sim d^\pi\\ a\sim\pi^{\mathrm{wc}}}}[A^\pi(s,a)]
\end{aligned}$$

For the second term:

$$
\begin{aligned}
\frac{2\gamma\epsilon_{V^\pi}^{\pi^{\rm wc}}}{(1-\gamma)^2}\mathop{\mathbb{E}}_{s\sim d^\pi}[D_{TV}(\pi\|\pi^{\rm wc})[s]] &= \frac{2\gamma\max_s|\mathbb{E}_{a\sim\pi^{\rm wc},\, s'\sim P}[\delta_{V^\pi}(s,a,s')]|}{(1-\gamma)^2}\mathop{\mathbb{E}}_{s\sim d^\pi}[D_{TV}(\pi\|\pi^{\rm wc})[s]] \\
&= \frac{2\gamma\max_s|\sum_a \pi^{\rm wc}(a|s)\sum_{s'}P(s'|s,a)\delta_{V^\pi}(s,a,s')|}{(1-\gamma)^2}\mathop{\mathbb{E}}_{s\sim d^\pi}[D_{TV}(\pi\|\pi^{\rm wc})[s]] \\
&= \frac{2\gamma\max_s|\sum_a \pi^{\rm wc}(a|s)A^\pi(s,a)|}{(1-\gamma)^2}\mathop{\mathbb{E}}_{s\sim d^\pi}[D_{TV}(\pi\|\pi^{\rm wc})[s]] \\
&\leq \frac{2\gamma\max_{s,a}|A^\pi(s,a)|}{(1-\gamma)^2}\mathbb{E}_{s\sim d^\pi}[D_{TV}(\pi\|\pi^{\rm wc})[s]]
\end{aligned}
$$

Therefore, we get the following bound:

**Radial-Inspired Bound**

$$
J(\pi)-J(\pi^{\rm wc}) \leq \frac{-1}{1-\gamma}\mathop{\mathbb{E}}_{\substack{s\sim d^\pi \\ a\sim\pi^{\rm wc}}}[A^\pi(s,a)] + \frac{2\gamma\max_{s,a}|A^\pi(s,a)|}{(1-\gamma)^2}\mathbb{E}_{s\sim d^\pi}[D_{TV}(\pi\|\pi^{\rm wc})[s]] \tag{16}
$$

To obtain an upper bound that can be estimated from samples collected under $\pi_k$, we follow the same steps as for SA-Reg, yielding the following upper bound:

$$
J(\pi)-J(\pi^{\rm wc}) \leq \frac{-1}{1-\gamma}\mathop{\mathbb{E}}_{\substack{s\sim d^{\pi_k} \\ a\sim\pi^{\rm wc}}}[A^{\pi_k}(s,a)] + (\alpha^A+\alpha^R)\mathop{\mathbb{E}}_{s\sim d^{\pi_k}}[D_{TV}(\pi_k\|\pi)[s]] + \alpha^A \mathop{\mathbb{E}}_{s\sim d^{\pi_k}}[D_{TV}(\pi\|\pi^{\rm wc})[s]] \tag{17}
$$

Where $\alpha^R = \frac{2\max_{s,a,s'}|R(s,a,s')|}{(1-\gamma)^2}$ and $\alpha^A = \frac{2\gamma\max_{s,a}|A^\pi(s,a)|}{(1-\gamma)^2}$

### B.3 Proof of Theorem 4: WocaR-RL Inspired Bound

In this subsection we provide a theoretical justification of the robustness loss used in WocaR-RL (Liang et al. (2022)) and why it needs SA-Reg.

We use Theorem 1, with the following function $\forall s\in S: f(s) = -V_{\rm wc}^\pi(s)$ and $\pi' = \pi^{\rm wc}$. We have the following:

$$
J(\pi)-J(\pi^{\rm wc}) \leq \frac{-1}{1-\gamma}L_{\pi,-V_{\rm wc}^\pi(s)}(\pi^{\rm wc}) + \frac{2\gamma\epsilon_{V^{\pi^{\rm wc}}}^{\pi^{\rm wc}}}{(1-\gamma)^2}\mathbb{E}_{s\sim d^\pi}[D_{TV}(\pi\|\pi^{\rm wc})[s]]
$$

For the first term:

$$
\begin{aligned}
\frac{-1}{1-\gamma}L_{\pi,-V_{\rm wc}^\pi(s)}(\pi^{\rm wc}) &= \frac{-1}{1-\gamma}\mathbb{E}_{\substack{s\sim d^\pi \\ a\sim\pi \\ s'\sim P}}\left[\left(\frac{\pi^{\rm wc}(a|s)}{\pi(a|s)}-1\right)\delta_{-V_{\rm wc}^\pi(s)}(s,a,s')\right] \\
&= \frac{-1}{1-\gamma}\mathbb{E}_{\substack{s\sim d^\pi \\ a\sim\pi \\ s'\sim P}}\left[\left(\frac{\pi^{\rm wc}(a|s)}{\pi(a|s)}-1\right)((R(s,a,s')-\gamma V_{\rm wc}^\pi(s')+V_{\rm wc}^\pi(s)))\right] \\
&= \underbrace{\frac{-1}{1-\gamma}\mathbb{E}_{\substack{s\sim d^\pi \\ a\sim\pi \\ s'\sim P}}\left[\left(\frac{\pi^{\rm wc}(a|s)}{\pi(a|s)}-1\right)R(s,a,s')\right]}_{L_1} + \underbrace{\frac{1}{1-\gamma}\mathbb{E}_{\substack{s\sim d^\pi \\ a\sim\pi \\ s'\sim P}}\left[\left(\frac{\pi^{\rm wc}(a|s)}{\pi(a|s)}-1\right)(\gamma V_{\rm wc}^\pi(s')-V_{\rm wc}^\pi(s))\right]}_{L_2}
\end{aligned}
$$

The $L_1$ term is the same as $\frac{-1}{1-\gamma}L_{\pi,0}(\pi^{\rm wc})$ used in the proof of SA-Reg. We focus now on the $L_2$:

$$
\begin{aligned}
L_2 &= \frac{1}{1-\gamma}\mathbb{E}_{\substack{s\sim d^\pi\\ a\sim\pi\\ s'\sim P}}\left[\left(\frac{\pi^{\text{wc}}(a|s)}{\pi(a|s)}-1\right)\left(\gamma V^\pi_{\text{wc}}(s')-V^\pi_{\text{wc}}(s)\right)\right]\\
&= \frac{1}{1-\gamma}\mathbb{E}_{\substack{s\sim d^\pi\\ a\sim\pi\\ s'\sim P}}\left[\left(\frac{\pi^{\text{wc}}(a|s)}{\pi(a|s)}-1\right)\left(R(s,a,s')-R(s,a,s')+\gamma V^\pi_{\text{wc}}(s')-V^\pi_{\text{wc}}(s)\right)\right]\\
&= \frac{-1}{1-\gamma}\mathbb{E}_{\substack{s\sim d^\pi\\ a\sim\pi\\ s'\sim P}}\left[\left(\frac{\pi^{\text{wc}}(a|s)}{\pi(a|s)}-1\right)R(s,a,s')\right]+\frac{1}{1-\gamma}\mathbb{E}_{\substack{s\sim d^\pi\\ a\sim\pi\\ s'\sim P}}\left[\left(\frac{\pi^{\text{wc}}(a|s)}{\pi(a|s)}-1\right)\left(R(s,a,s')+\gamma V^\pi_{\text{wc}}(s')-V^\pi_{\text{wc}}(s)\right)\right]\\
&\leq \frac{2\max_{s,a,s'}|R(s,a,s')|}{1-\gamma}\mathbb{E}_{s\sim d^\pi}[D_{TV}(\pi\|\pi^{\text{wc}})[s]]+\frac{1}{1-\gamma}\sum_s d^\pi(s)\sum_a\left(\pi^{\text{wc}}(a|s)-\pi(a|s)\right)A^\pi_{\text{wc}}(s,a)\\
&\leq \frac{2\max_{s,a,s'}|R(s,a,s')|}{1-\gamma}\mathbb{E}_{s\sim d^\pi}[D_{TV}(\pi\|\pi^{\text{wc}})[s]]+\frac{-1}{1-\gamma}\sum_s d^\pi(s)\sum_a \pi(a|s)A^\pi_{\text{wc}}(s,a)\\
&\leq \frac{2\max_{s,a,s'}|R(s,a,s')|}{1-\gamma}\mathop{\mathbb{E}}_{s\sim d^\pi}[D_{TV}(\pi\|\pi^{\text{wc}})[s]]+\frac{-1}{1-\gamma}\mathop{\mathbb{E}}_{\substack{s\sim d^\pi\\ a\sim\pi}}[A^\pi_{\text{wc}}(s,a)]
\end{aligned}
$$

Therefore, the upper bound of the first term is :

$$
\frac{-1}{1-\gamma}L_{\pi,-V^\pi_{\text{wc}}(s)}(\pi^{\text{wc}})\leq\frac{-1}{1-\gamma}\mathop{\mathbb{E}}_{\substack{s\sim d^\pi\\ a\sim\pi}}[A^\pi_{\text{wc}}(s,a)]+\frac{4\max_{s,a,s'}|R(s,a,s')|}{1-\gamma}\mathop{\mathbb{E}}_{s\sim d^\pi}[D_{TV}(\pi\|\pi^{\text{wc}})[s]]
$$

For the second term:

$$
\begin{aligned}
\epsilon^{\pi^{\text{wc}}}_{-V^\pi_{\text{wc}}(s)} &= \max_s\left|\mathbb{E}_{a\sim\pi^{\text{wc}},\,s'\sim P}\left[\delta_{-V^\pi_{\text{wc}}(s)}(s,a,s')\right]\right|\\
&= \max_s|\mathbb{E}_{a\sim\pi^{\text{wc}},\,s'\sim P}[R(s,a,s')+R(s,a,s')-(R(s,a,s')+\gamma V^\pi_{\text{wc}}(s')-V^\pi_{\text{wc}}(s))]|\\
&\leq \max_s|\mathbb{E}_{a\sim\pi^{\text{wc}},\,s'\sim P}[R(s,a,s')+R(s,a,s')]|+\max_s|\mathbb{E}_{a\sim\pi^{\text{wc}},\,s'\sim P}[(R(s,a,s')+\gamma V^\pi_{\text{wc}}(s')-V^\pi_{\text{wc}}(s))]|\\
&\leq 2\max_{s,a,s'}|R(s,a,s')|+\max_s|\mathbb{E}_{a\sim\pi^{\text{wc}},\,s'\sim P}[(R(s,a,s')+\gamma V^\pi_{\text{wc}}(s')-V^\pi_{\text{wc}}(s))]|\\
&\leq 2\max_{s,a,s'}|R(s,a,s')|+\max_s\underbrace{|\mathbb{E}_{a\sim\pi^{\text{wc}}}[A^\pi_{\text{wc}}(s,a)]|}_{=0}\\
&\leq 2\max_{s,a,s'}|R(s,a,s')|
\end{aligned}
$$

Therefore:

$$
\frac{2\gamma\epsilon^{\pi^{\text{wc}}}_{-V^{\pi^{\text{wc}}}}}{(1-\gamma)^2}\mathop{\mathbb{E}}_{s\sim d^\pi}[D_{TV}(\pi\|\pi^{\text{wc}})[s]]\leq\frac{4\gamma\max_{s,a,s'}|R(s,a,s')|}{(1-\gamma)^2}\mathop{\mathbb{E}}_{s\sim d^\pi}[D_{TV}(\pi\|\pi^{\text{wc}})[s]]
$$

Therefore, we get the following bound:

**WocaR-RL-Inspired bound**

$$
J(\pi)-J(\pi^{\text{wc}})\leq-\frac{1}{1-\gamma}\mathop{\mathbb{E}}_{\substack{s\sim d^\pi\\ a\sim\pi}}[A^\pi_{\text{wc}}(s,a)]+\frac{4\max_{s,a,s'}|R(s,a,s')|}{(1-\gamma)^2}\mathop{\mathbb{E}}_{s\sim d^\pi}[D_{TV}(\pi\|\pi^{\text{wc}})[s]] \tag{18}
$$

An upper bound that can be estimated from samples collected under $\pi_k$ is given by:

$$
J(\pi)-J(\pi^{\text{wc}})\leq-\frac{1}{1-\gamma}\mathop{\mathbb{E}}_{\substack{s\sim d^{\pi_k}\\ a\sim\pi}}[A^{\pi_k}_{\text{wc}}(s,a)]+3\alpha^R\mathop{\mathbb{E}}_{s\sim d^{\pi_k}}[D_{TV}(\pi_k\|\pi)[s]]+2\alpha^R\mathop{\mathbb{E}}_{s\sim d^{\pi_k}}[D_{TV}(\pi\|\pi^{\text{wc}})[s]] \tag{19}
$$

# C Experiment details

## C.1 Adversarial Training

All the agents are trained for 2 million steps. We use the same network architecture across all the algorithms and environments: MLP network with 2 layers, each with 64 neurons. We used the Adam optimizer.

For SA-Reg, the grid search considers robustness coefficient $\kappa$ in $\{0.001, 0.01, 0.05, 0.1\}$ ,actor and critic learning rate in $\{0.0001, 0.00025, 0.0004\}$ and entropy coefficient in $\{0, 0.001\}$. We use the same values for Radial. For WocaR-RL, we consider the actor, the critic, and the worst-case Q-network learning rates in $\{0.0001, 0.00025, 0.0004\}$, robust coefficient in $\{0.1, 0.5, 0.8\}$, soft polyak coefficient in $\{0.001, 0.05\}$ and entropy coefficient in $\{0, 0.001\}$. For SA-Reg, RAdial and WocaR-RL, we use the same $\epsilon$-scheduler: we slowly increase $\epsilon_t$ from zero until it reaches the full attack budget $\epsilon$ after 75% of the total training steps

For RoC-Mono, RoC-SA and RoC-Radial, we use actor and critic learning rates in $\{0.0001, 0.00025, 0.0004\}$, lagrangian learning rates in $\{0.00001, 0.0001, 0.001, 0.01, 0.05\}$, lagrangian maximum values in $\{1, 0.5, 0.1, 0.05\}$. For the RoC algorithms, we do not use an $\epsilon$-scheduler.

## C.2 Adversarial evaluation

We perform a grid search over the hyperparameters of the learned adversaries. Table 1 summarizes the search spaces considered for Robust Sarsa (RS), SA-RL and PA-AD. SA-RL and PA-AD adversary are trained for 2 million steps using PPO. For RS attacks, we train the for 200000 steps where the robust coefficient is zero, otherwise we train for 1 million step.

| Method | Hyperparameter | Values |
|---|---|---|
| RS | Learning rate | {1e-4, 2.5e-4, 5e-4, 1e-3, 2.5e-3, 5e-3} |
| | Activations | {relu, tanh} |
| | Robust coefficients | {0, 0.001, 0.01, 0.1, 0.5, 1} |
| | Polyak updates | {0.001, 0.005, 0.01, 0.05} |
| SA-RL | Actor learning rate | {1e-4, 5e-4, 1e-3, 2.5e-3, 5e-3} |
| | Critic learning rate | {1e-4, 5e-4, 1e-3, 2.5e-3, 5e-3} |
| | Entropy coefficient | {0, 1e-5} |
| | Squash actions | {True, False} |
| | Initial standard deviation | $\{-1, -2\}$ |
| PA-AD | Actor learning rate | {1e-4, 5e-4, 1e-3, 2.5e-3, 5e-3, 1e-2} |
| | Critic learning rate | {1e-4, 5e-4, 1e-3, 2.5e-3, 5e-3, 1e-2} |
| | Activation | {relu, tanh} |
| | Entropy coefficient | {0, 1e-3, 1e-4} |
| | Clip range | {0.2, 0.3} |

Table 1: Hyperparameter search spaces for learned attacks.



\end{document}